%% file: main.tex
\documentclass[letterpaper, 10 pt, conference]{ieeeconf}

\IEEEoverridecommandlockouts   %
\newif\ifANON
\ANONfalse

\input{preamble}

\title{\LARGE \bf NSDL: Neuro-Symbolic Division of Labor for Multi-Robot Task Planning}

\ifANON
  \author{Anonymous Author(s)}
\else
  \author{Xincheng Tan$^{1}$, Ke Wang$^{1}$, Yining Di$^{2}$, and Minfeng Zhu$^{1}$%
  \thanks{$^{1}$Xincheng Tan, Ke Wang, and Minfeng Zhu are with Zhejiang University, Hangzhou, China.
    Correspondence: {\tt\small minfeng\_zhu@zju.edu.cn}}%
  \thanks{$^{2}$Yining Di is with The Hong Kong Polytechnic University, Hong Kong SAR, China}%
  }
\fi

\begin{document}
\ifANON\else
  \AddToHookNext{shipout/foreground}{%
    \put(0.5\paperwidth,-\paperheight+30pt){\makebox(0,0)[b]{\parbox[b]{\textwidth}{\centering\footnotesize
      This work has been submitted to the IEEE for possible publication.\\
      Copyright may be transferred without notice, after which this version may no longer be accessible.}}}}
\fi
\maketitle
\thispagestyle{empty}
\pagestyle{empty}

\begin{abstract}
\input{sections/00_abstract}
\end{abstract}

\input{sections/01_intro}
\input{sections/02_related}
\input{sections/03_nsdl}
\input{sections/04_locus}
\input{sections/05_experiments}
\input{sections/06_conclusion}

\bibliographystyle{IEEEtran}
\bibliography{refs}

\clearpage
\input{sections/07_appendix}

\end{document}

%% file: preamble.tex
\usepackage{graphicx}
\usepackage{amsmath,amssymb,amsfonts}
\usepackage{booktabs}
\usepackage{multirow}
\usepackage{xcolor}
\usepackage{algorithm}
\usepackage{algpseudocode}
\usepackage{microtype}
\usepackage{stfloats}
\usepackage{xspace}

\usepackage{newtxtext}
\usepackage{newtxmath}

\usepackage{cite}
\usepackage{url}

\newcommand{\detect}{\texttt{detect}\xspace}
\newcommand{\sched}{\texttt{schedule}\xspace}

\AtBeginDocument{%
  \setlength\abovedisplayskip{4pt plus 2pt minus 2pt}%
  \setlength\belowdisplayskip{4pt plus 2pt minus 2pt}%
  \setlength\abovedisplayshortskip{2pt plus 1pt}%
  \setlength\belowdisplayshortskip{2pt plus 1pt}%
}

%% file: sections/00_abstract.tex
Planning for a heterogeneous robot team from a scene image and an
instruction involves two kinds of decisions. Some must be judged from
the scene: which goals are already met, how the task splits, and which
robot does what. Others follow from these and can be computed:
prerequisite actions, handoffs between robots, and timing. We argue
that each kind should go to the component suited to it: a small
vision-language model learns the first, and a deterministic solver
computes the second. In NSDL (Neuro-Symbolic Division of Labor), the
trained model writes its judgments as a plan sketch, which lists the
goals that still need action and the intended steps in order, each with
a robot. The solver expands the sketch into an executable joint plan.
The model calls two tools, a detector for evidence and the solver, and
is trained with the solver in the loop. We evaluate on the public
VIKI-L2 benchmark and on LOCUS-Bench, a new benchmark that grades the
two kinds of difficulty separately. NSDL outperforms end-to-end
planners and formalize-then-plan baselines on both benchmarks and with
three backbones from 3B to 8B; the 7B model reaches 97.0\% %
out-of-distribution accuracy on VIKI-L2. Ordering and scheduling errors
nearly disappear once the solver writes the structure; the remaining
failures are misjudged states, and most goals the model misses are
ones it never lists. Code, model weights, and LOCUS-Bench are publicly
available at \mbox{\url{https://github.com/Xincheng-Tan/NSDL}}.

%% file: sections/01_intro.tex
\section{INTRODUCTION}
\label{sec:intro}

Multi-robot systems combine robots with complementary capabilities for
tasks that no single robot completes alone. Planning for such a team
from a scene image and an instruction has become a task for
vision-language models (VLMs). The planner receives the instruction,
one image, and the capabilities of each robot, and returns a joint plan
that says which robot does what and in what order
\cite{korsah13,vikir}. Large language models (LLMs) brought broad
reasoning to this setting \cite{saycan,roco,lammap}, and VLMs connect it
to raw visual observation \cite{embodiedgpt,robobrain,vikir}. Trained
planners for this task build on small VLMs \cite{vikir}, whose capacity
is limited. Which decisions should such a model learn, and which should
a solver compute? The question applies to language-model planning in
general \cite{llmp,viplan}.\looseness=-1

\begin{figure}[t]
  \centering
  \includegraphics[width=\columnwidth]{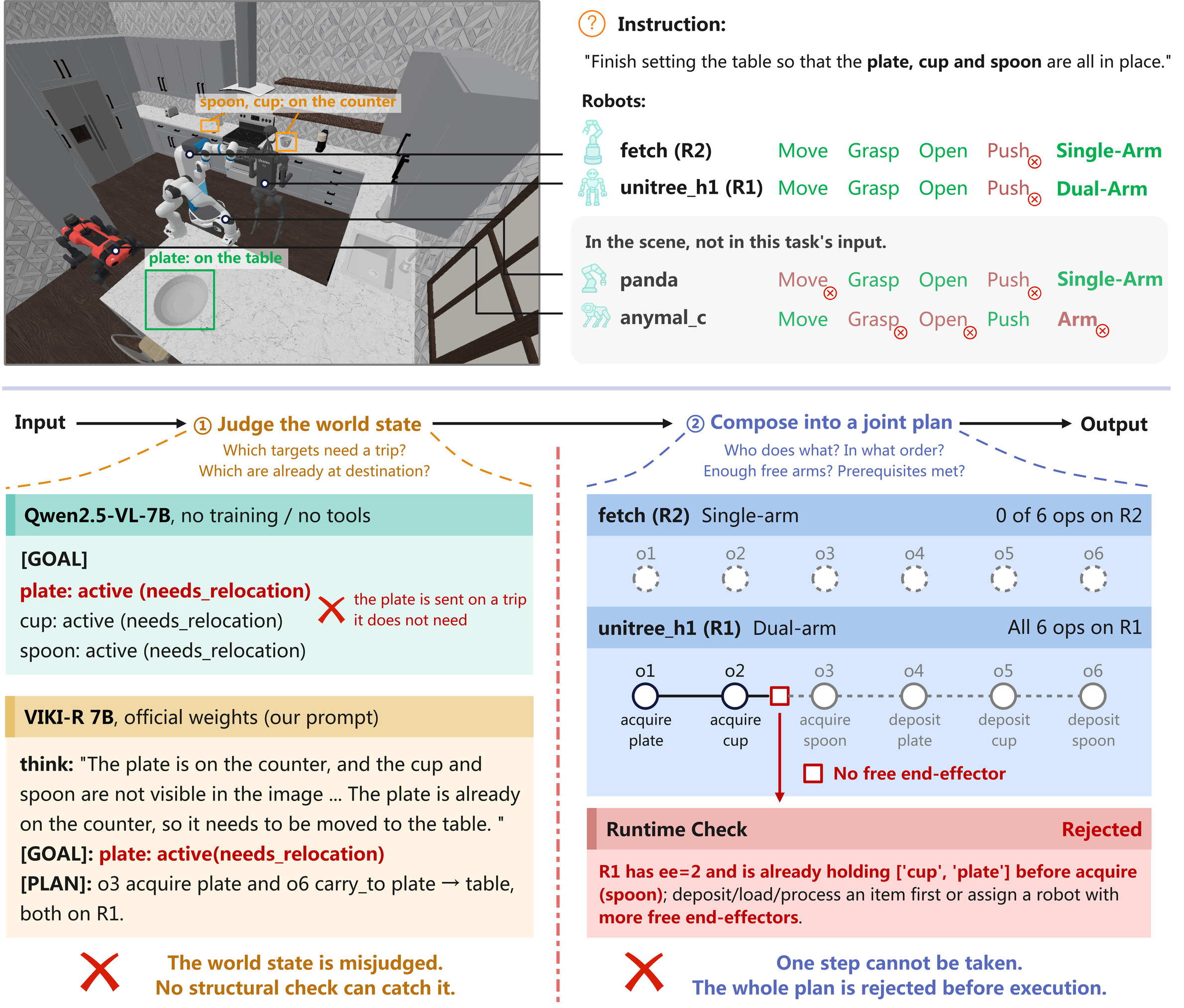}
  \vspace{-0.45cm}
  \caption{\textbf{One scene, two failures.} Left: both planners mark a
  plate that is already on the table as still to be moved. Right: the
  plan of the untrained Qwen2.5-VL-7B requests a third pick for a robot
  whose two hands are occupied, and the runtime rejects it. Sample from
  LOCUS-Bench (Sec.~\ref{sec:locus}).} %
  \label{fig:teaser}
\end{figure}

The decisions in this task fall into two kinds. \emph{State judgment}
covers whether each goal is already satisfied in the scene, how the
task decomposes into subtasks, and which robot takes which. No fixed
procedure computes these from the input; they require image evidence
and semantic understanding. \emph{Structural planning} covers the
prerequisites and the scheduling that follow once intent and assignment
are fixed, such as opening a cabinet before stowing an item and
respecting end-effector capacity. Given a fixed action model, a
deterministic solver computes them correctly. Fig.~\ref{fig:teaser}
shows one failure of each kind on one scene, made by two planners: the
untrained Qwen2.5-VL-7B and VIKI-R 7B, an end-to-end planner trained on
the public VIKI-Bench. Both mark a plate that is already on the table as
still to be moved, and both stack picks onto a robot without a free
end-effector.\looseness=-1 %

We argue that each kind should go to the component suited to it: the
model learns the state judgment, and a deterministic solver computes
the structural planning. The model then spends its capacity only on the
decisions that need the scene, and a 7B model plans better than
planners that learn both kinds. Existing planners draw the line
elsewhere: end-to-end planners \cite{vikir,partnr,embodiedgpt} learn
both kinds, and formalize-then-plan methods
\cite{lammap,osdag,imrllm,vilain,vlm-formalizer} have an untrained
model write the input of a classical planner, which then produces the
plan (Sec.~\ref{sec:related}).\looseness=-1

NSDL (Neuro-Symbolic Division of Labor) implements this division
(Fig.~\ref{fig:overview}). The trained model writes its judgments as a
plan sketch: the goals that still need action and the intended
operations (ops) in order, each assigned to a robot. A deterministic
solver $\Phi$ expands the sketch into the executable plan: it adds
prerequisite actions and handoffs, breaks each op into primitive
actions, and schedules them. The model works in one multi-turn dialogue
(a trajectory) with two tools. \detect returns the objects and spatial
relations it observes, and \sched runs the solver; when the solver
rejects a sketch, \sched returns the reason, and the model revises the
sketch. The model is trained by supervised fine-tuning and then by
reinforcement learning with the solver in the loop.\looseness=-1

We evaluate NSDL on the public VIKI-L2 benchmark, the task-planning
level of VIKI-Bench \cite{vikir}, and on LOCUS-Bench, which we build
because public benchmarks fold both kinds of failure into one score.
LOCUS-Bench sets state-judgment and structural-planning difficulty on
two separate five-level scales, and its out-of-distribution splits
raise the difficulty of one kind, the other, or both. On both
benchmarks and with three backbones from 3B to 8B, NSDL outperforms
end-to-end planners and formalize-then-plan baselines; the 7B model
reaches 97.0\% %
out-of-distribution accuracy on VIKI-L2.\looseness=-1

Our contributions are as follows.
\begin{itemize}\setlength\itemsep{2pt}\setlength\parskip{0pt}
\item \textbf{NSDL.} A framework for embodied multi-robot task
planning in which a trained model makes the decisions that depend on
the scene and writes them as a plan sketch, and a deterministic solver
expands the sketch into the executable joint plan.
\item \textbf{LOCUS-Bench.} A benchmark that sets state-judgment and
structural-planning difficulty on separate five-level scales, so that
a failure can be traced to either kind.
\item \textbf{Findings.} Once the solver writes the structure, the
runtime almost never rejects a plan for its ordering or scheduling, and
these errors return when the model is trained without the solver. The remaining failures are misjudged
states. Most goals the model misses are ones it never lists rather
than ones it misreads, and ground-truth evidence does not reduce their
number.\looseness=-1
\end{itemize}

%% file: sections/02_related.tex
\section{RELATED WORK}
\label{sec:related}

Classical task planning and multi-robot task allocation compute correct
plans once the symbolic input is given, and their limitation is the
input itself \cite{pddl,fastdownward,korsah13}: the world state, the
goals, and the capabilities must be written by hand in symbolic form.
Our solver plays this role with a narrower scope, completing,
decomposing, and scheduling a given sketch (Sec.~\ref{sec:method});
our question is which component should turn the image and the
instruction into the input it needs.\looseness=-1

The end-to-end line entrusts the entire plan to the model. Prompt-based
methods expand instructions into grounded skill sequences
\cite{saycan}, and multi-robot decomposition, allocation, and
negotiation are carried out by the LLM \cite{roco,coela}. A trained
branch places perception and planning in the weights of a (V)LM
\cite{embodiedgpt,robobrain,partnr}. Within this branch, VIKI-R
\cite{vikir} trains a small VLM with chain-of-thought supervised
fine-tuning (SFT) and GRPO \cite{grpo} on a heterogeneous multi-robot
benchmark and is our most direct comparison. We share the trained model
that reads the scene and differ on where the structure is produced.\looseness=-1

Formalize-then-plan methods let the model formalize the task for a
classical planner; the model's role in this line is called
LLM-as-formalizer \cite{llmformalizer}. The task is formalized from
text into the Planning Domain Definition Language (PDDL) \cite{llmp}, from images by an untrained
detector-plus-LLM or a prompted VLM
\cite{vilain,vilain-tamp,vlm-formalizer}, or by multi-robot front ends
that write per-robot problems or dependency graphs
\cite{lammap,osdag,imrllm}. The text and multi-robot branches receive
the state symbolically. The image branch evaluates the model that reads
the scene but does not train it, and \cite{vlm-formalizer} locates the
bottleneck in initial-state relations. Correction loops fire on syntax
errors, undefined objects, contradictions, unreachable goals, or
planning failures \cite{vilain,vilain-tamp}. A wrong state that is
internally consistent fires none of them, and per-action re-checks keep
the judgment in an untrained model \cite{dkprompt}.\looseness=-1

Because such errors admit no downstream correction, the accuracy of
this judgment is the only lever; across this line the judgment that
formalizes the state is untrained, and we train it. %
VLMFP \cite{vlmfp} fine-tunes a VLM that describes the initial scene
and simulates action consequences in natural language, while an
untrained model formalizes the scene into the initial state. %
In the dichotomy of \cite{viplan} between VLM-as-planner and
VLM-as-grounder, we take the combination that neither occupies: state
judgment in a trained model, structural planning in a deterministic
solver, within one trajectory the model is trained on.\looseness=-1

Benchmarks in this family report a single aggregate score
\cite{coela,roco,partnr,lammap,vikir}. EAI \cite{eai} decomposes
evaluation by interface module and PRISM \cite{prism} grades tasks by
capability with swappable oracle perception probes, but neither sets
state judgment and structural planning as separately given dimensions
at construction time, so their attribution cannot distinguish the two.
LOCUS-Bench sets the two at construction time
(Sec.~\ref{sec:locus}), with oracle probes following \cite{prism},
which we call ground-truth rows; VIKI-L2 (VIKI for short) serves as our
public external yardstick.\looseness=-1

%% file: sections/03_nsdl.tex
\section{NSDL}
\label{sec:method}

\begin{figure*}[t]
  \centering
  \includegraphics[width=\textwidth]{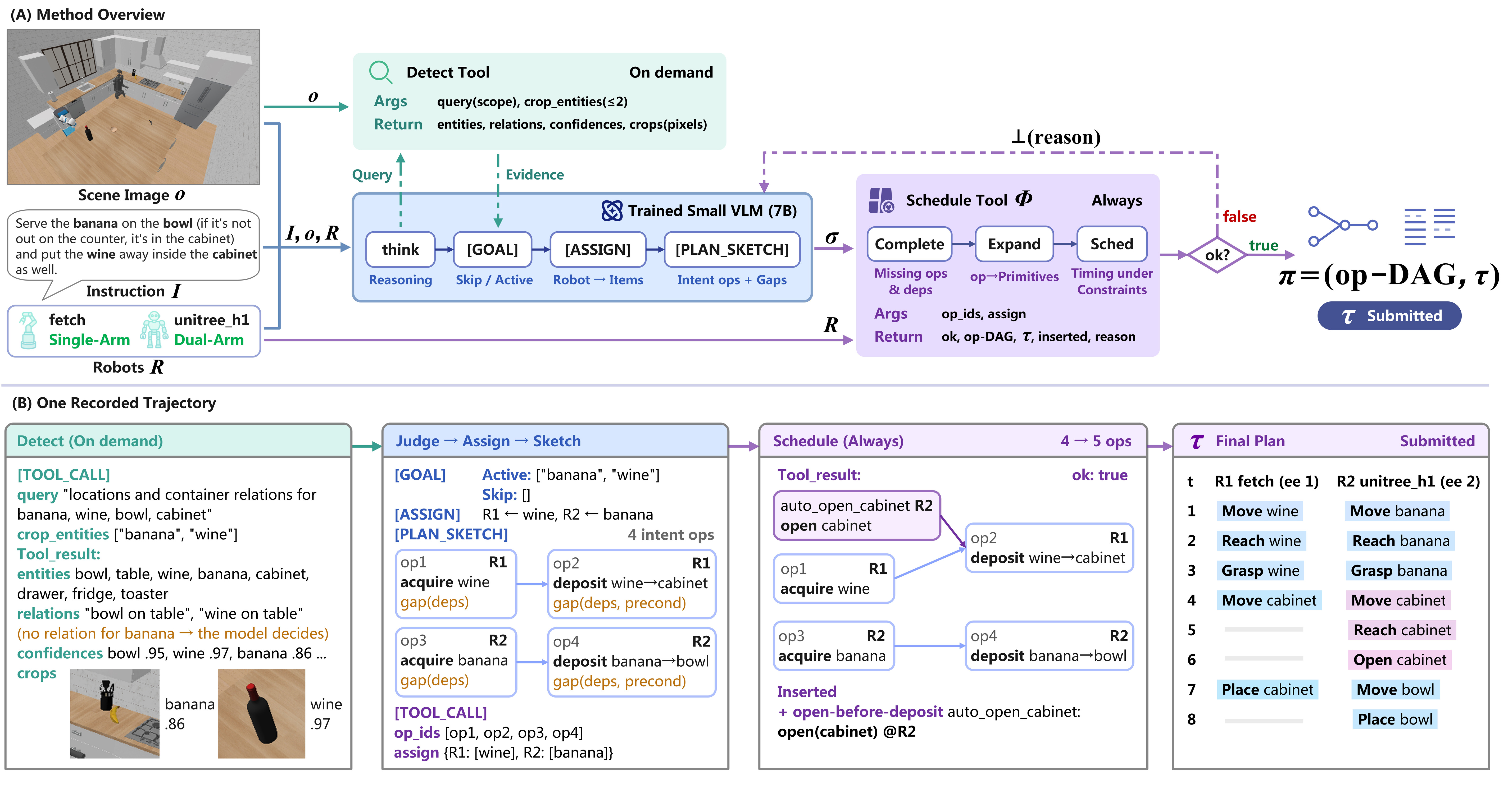}
  \vspace{-1.00cm}
  \caption{\textbf{NSDL overview (A) and one recorded trajectory (B).}
  (A) The trained model writes the goals that still need action, the
  assignment, and the intended steps as a plan sketch $\sigma$; the
  solver $\Phi$ expands $\sigma$ into the plan $\pi$ with its
  dependency graph and schedule $\tau$, or rejects it with a structured
  reason; \detect returns observed entities and relations and never
  answers task questions. (B) LOCUS sample 00001, NSDL 7B: the sketch
  names no \texttt{open}; $\Phi$ inserts one (highlighted) and
  schedules both robots in parallel.}
  \label{fig:overview}
  \vspace{-0.10cm}
\end{figure*}

NSDL lets a trained model make every state-judgment decision inside one
agentic trajectory and hands the resulting plan sketch to a
deterministic solver that expands it into the submitted plan
(Fig.~\ref{fig:overview}(A)). The model reads the instruction, the
image, and the robot team and writes, in a fixed order after a think
segment, \texttt{[GOAL]} (a \texttt{skip} or \texttt{active} verdict
for each goal with a one-sentence reason), \texttt{[ASSIGN]}, and
\texttt{[PLAN\_SKETCH]}, calling \detect first in a separate turn when
it needs evidence. A finished sketch is always handed to \sched, which
runs the solver $\Phi$ and returns the expanded plan or a rejection
with a structured reason; once $\Phi$ accepts, the expansion is the
submitted plan. This section defines the problem and the sketch
(Sec.~\ref{sec:problem}), the trajectory and its two tools
(Sec.~\ref{sec:method-traj}), the solver (Sec.~\ref{sec:method-sketch}),
the detector (Sec.~\ref{sec:method-detect}), and the training recipe
(Sec.~\ref{sec:method-training}).\looseness=-1

\subsection{Problem and the Plan Sketch}
\label{sec:problem}

A task instance gives an instruction, one scene image, and a robot team
with capability labels and end-effector counts, and no symbolic world
state: $x = (I, o, R)$. Executability is defined by the action model
$\mathcal{M}$ (end-effector capacity, holding, closed containers
isolating their contents, per-step concurrency) over the eight
primitives $\mathcal{A} = \{$Move, Reach, Grasp, Place, Open, Close, Push,
Interact$\}$ of the official VIKI-R implementation, shared by both
benchmarks; we re-encode $\mathcal{M}$ in STRIPS style with a typed
entity catalog $E$, used by $\Phi$ and by our runtime validator
(one difference, a Move demanded before each deposit, is described in
Appendix~\ref{app:failure}). The output
$\pi = (O, \rho, \prec, \tau)$, comprising ops, an assignment, a
directed acyclic dependency graph, and a primitive schedule, is scored through $\tau$
against a ground-truth initial state $s_0$ (of which $o$ is the
rendering) and a reference plan $\bar{\tau}$ visible only to scoring:
\begin{equation}
\label{eq:sol}
\begin{split}
\mathrm{Sol}(\pi; x, s_0, \bar{\tau}) :\Leftrightarrow\;
& \mathrm{Exec}_{\mathcal{M}}(\pi, s_0) \\
& \wedge\, s_T(\pi, s_0) \models G(I, s_0)
\,\wedge\, |\tau| \le |\bar{\tau}|
\end{split}
\end{equation}
(i) executable under $\mathcal{M}$ from $s_0$; (ii) goal-achieving,
where $G$ itself depends on $s_0$, since which goals require action
varies with the scene; and (iii) within the reference length.
Structural errors break (i); judgment errors break (ii) or (iii), and a
sketch whose asserted states, entities, or assignments mismatch the
scene breaks (i). Scoring is symbolic replay only.\looseness=-1

The state judgment is written into the plan sketch, and the solver
expands it without ever reading the image or the verdicts. The sketch
is
\begin{equation}
\label{eq:sketch}
\sigma = (g, \rho_\sigma, O_\sigma), \;
O_\sigma \in (\mathcal{V}_\sigma {\times} E {\times} (E \cup \{\varnothing\})
{\times} (R \cup \{\varnothing\}))^{*}
\end{equation}
where each op carries its type, entity, destination, and robot; the
verdict $g$ marks each goal \texttt{skip} or \texttt{active} over the
candidate goals the model lists from $I$ and $o$ (a goal the model
never lists is an enumeration error, which counts as a judgment error); the partial
assignment $\rho_\sigma$ fills the ops whose robot field is blank; and
$O_\sigma$ is a sequence of intended steps over the sketch vocabulary
$\mathcal{V}_\sigma$ (Sec.~\ref{sec:method-sketch}), with deterministic
details omissible and no timing; its order is the only ordering
information the model gives. The solver is a map
$\Phi: (\sigma, E, R) \mapsto \pi$ or $\bot$ with a structured reason;
it reads no image and never reads $g$, a skipped goal reaching it only
as the absence of its ops. Three properties hold by construction
(Sec.~\ref{sec:method-sketch}). $\Phi$ is a function, so a given
sketch always yields the same expansion. An accepted expansion is executable, under our
encoding, in the state the sketch asserts (the states its ops
presuppose, the catalog types, and fixed defaults: containers closed,
end-effectors free); when these assertions do not match the scene, the
accepted plan may be inexecutable under $s_0$. And $\Phi$ preserves
$g$, the fields of every op, and the given ordering verbatim, adding
only mechanical ops and dependency edges, so judgment errors pass
through and $\Phi$ cannot repair them. We learn
the front end $f_\theta: (I, o, R, \varepsilon_{1:k}) \mapsto \sigma$
that maximizes $\mathrm{Sol}(\Phi(\sigma))$ (Eq.~(\ref{eq:objective})),
with evidence $\varepsilon$ requested on demand from \detect.\looseness=-1

\subsection{The Trajectory and Its Tools}
\label{sec:method-traj}

Each assistant turn issues at most one tool call, and the two tools
split the work by what they are allowed to return. \detect takes a
query naming the entities the model is uncertain about, plus up to two
of them for close-up crops, and returns the entities present,
fixture-anchored relations, confidences, and the crops. \sched takes
references to sketch ops plus an assignment completing unwritten robot
fields, and returns the completed op-level DAG, the schedule $\tau$,
and an itemized list of insertions, or a structured error code; the
model writes neither primitives nor timing. Upon rejection, the model
rewrites \texttt{[ASSIGN]} and \texttt{[PLAN\_SKETCH]} (never
\texttt{[GOAL]}) and calls \sched again; the submitted $\tau$ is the
last accepted schedule (fallback protocol and counts in
Appendix~\ref{app:protocol}). \detect answers observation questions
(\texttt{apple on table}) and never task questions (whether the apple
is already at its destination), which remain with the model; whether to
call it and what to query are learned, and nothing beyond the turn
budget and the one-call rule is enforced. \sched does all of the
structural planning.\looseness=-1

\subsection{The Solver}
\label{sec:method-sketch}

The plan is built on three tiers of vocabulary, and the sketch carries
only the tier that bears intent. The eight intent op types are acquire
and its in-place variant local\_acquire, deposit, open, process\_at,
apply\_tool, activate, and inspect; the op-DAG vocabulary
$\mathcal{V}_\pi$ adds six mechanical types (carry\_to, handoff,
load\_carrier, push\_carrier, unload\_carrier, and close); and the
scored $\tau$ is defined over the primitives $\mathcal{A}$ of
Sec.~\ref{sec:problem}. Ops inserted by $\Phi$ never add intent;
\texttt{open} plays a dual role, nameable as intent and insertable as a
prerequisite, as in Fig.~\ref{fig:overview}. On VIKI,
$\mathcal{V}_\sigma$ is the eight intent types, whereas on LOCUS some
mechanical ops written by the model carry intent themselves (the
intermediate destination of a carry\_to), so
$\mathcal{V}_\sigma = \mathcal{V}_\pi$.\looseness=-1

Collaboration is expressed by the assignment alone, and the solver
turns it into structure in three deterministic stages,
$\mathrm{SCHED} \circ \mathrm{EXPAND} \circ \mathrm{COMPLETE}$, run in
full by the \sched tool: assigning acquisition and placement of one
entity to different robots implies a cross-robot transfer, whose
concrete form $\Phi$ infers from feasibility. Stage 1 (COMPLETE), rule-based and
search-free, covers three structure classes: an open before the first
deposit into a closed openable container; a handoff when the depositor
of an entity is not its acquirer; and a carry\_to for apply\_tool.
Executors of inserted ops come from capability labels and end-effector
counts only, dependency edges from custody (which robot holds an entity),
open-before-deposit, and
carrier chains; COMPLETE deletes no op, synthesizes no intent op, and
never alters a model-written field. Stage 2 (EXPAND) decomposes each
op into primitives by textbook backward regression
\cite{pddl,fastdownward} in topological order: the start state is
assembled from the end states of predecessors, and regression is
applied locally to each op from its postconditions, with no global
planning.\looseness=-1

Stage 3 (SCHED) arranges the primitives by greedy
earliest-ready list scheduling under four hard constraints (precedence,
one primitive per robot per step, end-effector capacity, and effect
visibility, an effect becoming visible to the other robots from the
next step); interleavings follow from these
constraints, no step-minimality is claimed, and condition (iii) is
checked at scoring, where it matches the official length gate
(Appendix~\ref{app:protocol}). Global checks
reject dependency cycles, capability mismatches, multi-robot
acquisition, and carrier conflicts. Unlike PDDL-family planners, $\Phi$
receives no initial state and no goal, exactly what our model must
judge, and only completes, decomposes, and schedules; its pseudocode
and the label and predicate vocabulary of $E$ ship with the
code.\looseness=-1

\subsection{\detect}
\label{sec:method-detect}

\detect is backed by an in-domain YOLO detector, shared by every
tool-taking method we compare, and it returns evidence that the model
still has to read. The backend is a YOLO26-L detector \cite{yolo26},
one per benchmark, trained only on official train-split images (mAP50
.952 on the 560 LOCUS evaluation images). %
Relations are computed from detection boxes alone, reading neither the
instruction nor destinations; crops return task-blind pixels. The
detector misses two things by design: relations anchor on fixtures only, so a
goal destined for a movable container never produces a record (the
banana in Fig.~\ref{fig:overview}), and closed containers hide their
interior. An absent record therefore does not mean an absent object:
the tool never asserts absence, and reading an absent record is itself
a state judgment, left to the model. The natural alternative, judging
by rules (detected at the destination means skip), fails exactly at
these two gaps, forced to read a missing record as \emph{not
there}; this is how the formalize-then-plan baselines of
Sec.~\ref{sec:exp-setup}
fail.\looseness=-1

\subsection{Training}
\label{sec:method-training}

Training optimizes the front end with the solver fixed, first by
supervised fine-tuning on trajectories whose tool returns come from the
same solver used at test time:
\begin{equation}
\label{eq:objective}
\begin{split}
& \max_\theta \; \mathbb{E}_x\!\left[\, \mathrm{Sol}(\Phi(\sigma); x,
s_0, \bar{\tau}) \,\right], \\
& \sigma \sim f_\theta(\cdot \mid I, o, R, \varepsilon_{1:k}), \;\;
\varepsilon_i = \mathtt{detect}(q_i; o), \;\; q_i \sim f_\theta
\end{split}
\end{equation}
For SFT, \texttt{[GOAL]}, \texttt{[ASSIGN]}, and
\texttt{[PLAN\_SKETCH]} are derived backward from the reference plan
and the ground-truth initial state, then run through the real \sched to
backfill the tool interaction. \detect demonstrations are injected by
rule, with queries taken from instruction-named candidates, never from
the annotated answer. A hard check drops samples where a skip-judged
goal has an acquisition op, and rejection-and-repair turns are injected
synthetically on 3\% of samples. %
Each benchmark has its own corpus derived from its official train split
only (VIKI 7196 rows, LOCUS 6300), %
trained separately; think segments are synthesized by a closed-source
teacher (swapped in Sec.~\ref{sec:exp-judgment}). The model also writes the
expanded plan back as \texttt{[PLAN]}; the reward scores that copy, so
the model learns to carry the solver's plan faithfully
(Appendix~\ref{app:training}).\looseness=-1

A short reinforcement learning (RL) phase then refines the same model
with the solver in the loop. It follows tool-integrated practice
\cite{retool}, GRPO \cite{grpo} with DAPO stabilizations \cite{dapo}
(budget in Sec.~\ref{sec:exp-setup}, hyperparameters in
Appendix~\ref{app:training}). The reward
$\mathcal{R} = w_T R_{\mathrm{task}} + w_S R_{\mathrm{state}} + w_E
R_{\mathrm{ent}}$ reads only training-set annotations:
$R_{\mathrm{task}}$ is the official score of the \texttt{[PLAN]} the
model writes, re-expanded as in the fallback of
Appendix~\ref{app:protocol}; $R_{\mathrm{state}}$
and $R_{\mathrm{ent}}$ are Jaccard overlaps of the active-goal set and
the acquired-entity set against annotation; two small capped protocol
penalties subtract from the weighted sum. Evaluation submits the last
$\Phi$-accepted schedule, since only it carries the executability
property above.\looseness=-1

%% file: sections/04_locus.tex
\section{LOCUS-BENCH}
\label{sec:locus}

LOCUS-Bench (LOCUS for short) grades state-judgment and
structural-planning difficulty on two separate five-level scales and
crosses them into a grid; its name refers to the locus of competence
where a failure lands. Both scales are set at construction time from
the two kinds of decision.\looseness=-1

The state-judgment scale (S) asks progressively more of the model's
reading of the scene, and the structural-planning scale (M) asks
progressively more of the expansion. S0: no judgment. S1: whether a
single named target is already in place. S2: which of 2 to 3
conditional candidates are absent. S3: which of 3 to 6 quantified
instances are unsatisfied. S4: whether invisible implies absent under
occlusion, with single-frame fallback planning; observability is
guaranteed by construction, the target being either on a visible
tabletop or inside an opaque closed cabinet, and the instruction
declares the unique fallback. All added difficulty at S$\ge$1 is
carried by $g$ and the enumerated goal set of
Eq.~(\ref{eq:sketch}). M0: no structure. M1: an opening-prerequisite
chain. M2: independent subtasks decoupled across robots in parallel.
M3: coupled collaboration, the ops of one robot depending on those of
another (a shared opening, a forced handoff). M4: a forced relay chain,
an immobile robot requiring a pushable carrier. The structure of each
level is scored pass or fail by symbolic replay.\looseness=-1

Ten task families instantiate the grid, and the evaluation splits
extrapolate along one scale, the other, or both. The families
(transport, conditional fetching, occlusion fallback, carrier relay,
patrol, and five more released with the code) are rendered by the
official VIKI renderer over six heterogeneous robots with a 34-class
vocabulary (20 from VIKI, 14 new). %
Four combinations are absent (hatched in Fig.~\ref{fig:locus_grid}):
S4$\times$M0 is structurally impossible, since a hidden target forces
an open, and no family instantiates S2--S4 $\times$ M4, the relay
family spanning S0--S1; 21 cells remain. %
The sampling unit is the family-cell, one family in one cell; 49
family-cells instantiate the 21 cells. Train (6300) and the in-distribution (ID) test
split (180) cover the same family-cells within S0--S2 and M0--M2; the
out-of-distribution (OOD) splits extrapolate: OOD-S (140) only along S (S$\ge$3, M$\le$2),
OOD-M (200) only along M (M$\ge$3, S$\le$2), and OOD-X (40) along both
(S$\ge$3 at M3). %
Samples with an empty reference plan (73 across evaluation splits,
Table~\ref{tab:locus_stats}) all sit in rows S$\ge$1: the correct
output is no action at all.\looseness=-1

Instruction templates map to several ground-truth situations at
uniform frequency, so the text alone never decides the plan;
opposite-behavior situations share byte-identical text, and
candidate-destination pairs are fully crossed in the training grid,
so extrapolation adds no new lexical combination.\looseness=-1

Scoring reuses the official VIKI-L2 scorer and format on symbolic
replay, and every reference plan replays through the same simulator
that scores predictions. An empty-plan extension makes a prediction
correct on an empty reference only if it is itself empty. Under the
length gate, a plan longer than the reference scores 0, so re-moving
an already placed object fails unless the rest of the plan is shorter
than the reference. %
Appendix~\ref{app:protocol}.\looseness=-1

%% file: sections/05_experiments.tex
\section{EXPERIMENTS}
\label{sec:exp}

\subsection{Setup}
\label{sec:exp-setup}

One recipe and one tool interface serve both benchmarks, with a
separate corpus and one checkpoint each. The main backbone is
Qwen2.5-VL-7B, the backbone trained by VIKI-R; a same-family 3B model
probes capacity, and Qwen3-VL-8B probes backbone transfer. VIKI-L2 is
evaluated on its official splits of 1800 ID and 1218 OOD %
rows; the OOD split holds out whole task
families with conditional instructions, whose correct plan depends on
judging from the image which named objects are already in place. LOCUS
uses the four splits of Sec.~\ref{sec:locus}. Think segments are
synthesized by the closed-source \mbox{gpt-5.6-terra} \cite{gpt56} (the teacher
is swapped in Sec.~\ref{sec:exp-judgment}). RL runs 30 steps, %
with reward weights and hyperparameters in Appendix~\ref{app:training},
and all trained rows are single runs. We name the 7B checkpoints NSDL-SFT and
NSDL-RL. NSDL-RL, the full recipe, is the headline row on both
benchmarks; NSDL-SFT is the checkpoint that every ablation and every
paired test in Secs.~\ref{sec:exp-main} to \ref{sec:exp-judgment}
uses.\looseness=-1

Both benchmarks are scored by the official VIKI-L2 scorer on symbolic
replay, and we report alongside it the rate at which our runtime
validator rejects a submitted plan. LOCUS frames are capped at 524288
pixels %
for the VLM, while the detector reads native resolution. The tool-turn
budget is 3 executed tool calls, %
including $\Phi$-rejected \sched calls, with the final answer turn free;
decoding stops after an issued tool call (ablated in
Sec.~\ref{sec:exp-judgment}). The submitted plan is the last
$\Phi$-accepted schedule, with a fallback when $\Phi$ never accepts
(Appendix~\ref{app:protocol}). The rejection rate (rej.\ in the tables)
comes from a validator that rejects any plan it cannot verify and reads
no annotation; rejected rows score 0 and stay in the denominator.
$\Phi$-accepted schedules pass by construction, so for NSDL it arises
only on fallback rows, while for the variant without $\Phi$ it
measures rejected model-written structure; \mbox{--} marks rows where
the validator does not run. Comparisons use exact per-sample McNemar
tests with discordant counts $a/b$: rows solved only by the first-named
configuration over rows solved only by the second. %
Differences in the text are whole-set readings; table means weight the
splits equally.\looseness=-1

We compare NSDL with end-to-end planners and with formalize-then-plan
baselines built on our detector and solver, in which fixed rules or an
untrained model decide which goals are done. The
end-to-end rows on VIKI-L2 are VIKI-R, its answer-only SFT variant
(Ans-SFT), and zero-shot models, transcribed from \cite{vikir} (rep.).
On LOCUS, the released VIKI-R weights run under two protocols: our
op-level protocol (the \texttt{[GOAL]}/\texttt{[PLAN]} prompt and tool
interface of this paper) and their native primitive-schedule protocol
(their own prompt and resolution, scored by the official VIKI-L2
scorer); \mbox{gpt-5.6-terra} runs with our full toolset, and the transfer row
evaluates NSDL-SFT trained on VIKI. The three formalize-then-plan
baselines are prompt-only, where an untrained model judges the states
itself; full-translation, where the model only translates and fixed
rules give the verdicts (it reuses the NSDL-SFT checkpoint, which was
not trained to translate); and trained-translation, where
the model is trained on the same rows relabeled judgment-free, every
candidate translated and the verdicts left to the rules. Two settings
differ from NSDL's: on VIKI their vocabulary admits mechanical ops, and
a translation baseline that asks for evidence without writing a
sketch is told that none will be returned and gets one more
turn. %
In Tables~\ref{tab:viki} and \ref{tab:locus}, the Judgment column
names the component that gives the goal verdicts (the model, an
untrained model, fixed rules, or ground truth), and bold marks the best
entry among method rows, the variant without $\Phi$ excluded; the $\Phi$
column of Table~\ref{tab:locus} marks whether the solver expands the
plan. Table~\ref{tab:locus} groups its rows into A, no training on
LOCUS; B, trained on LOCUS with the goal verdicts given by fixed rules; C,
trained on LOCUS with the state judgment in the model; and D, the
reference row, ground-truth sketches expanded by $\Phi$.\looseness=-1

\subsection{Dividing the Labor Leads on Both Benchmarks}
\label{sec:exp-main}

On VIKI-L2 the same 7B backbone reaches near-ceiling accuracy in and
out of distribution under NSDL, while end-to-end training on the same
data collapses out of distribution (Table~\ref{tab:viki}). The 3B and
Qwen3-VL-8B backbones follow the same pattern. RL neither gains nor
loses on VIKI, where SFT is already near ceiling
($p=.45$). %
The formalize-then-plan baselines stay far below NSDL on both splits;
the last paragraph of this section examines why.\looseness=-1

\begin{table}[t]
\caption{VIKI-L2 accuracy (ID / OOD) and runtime rejection rate.
Judgment: the component that gives the goal verdicts.}
\label{tab:viki}
\centering
\small
\setlength{\tabcolsep}{3.0pt}
\resizebox{\columnwidth}{!}{%
\begin{tabular}{llccc}
\toprule
Method & Judgment & ID & OOD & rej.\% ID/OOD \\
\midrule
\multicolumn{5}{l}{\emph{End-to-end}} \\
Qwen2.5-VL-7B zero-shot (rep.) & untrained & .004 & .000 & -- \\
Qwen2.5-VL-72B zero-shot (rep.) & untrained & .084 & .012 & -- \\
GPT-4o zero-shot (rep.) & untrained & .226 & .100 & -- \\
VIKI-R 3B (rep., SFT+GRPO) \cite{vikir} & model & .936 & .321 & -- \\
Ans-SFT 7B (rep.) \cite{vikir} & model & .969 & .256 & -- \\
VIKI-R 7B (rep., SFT+GRPO) \cite{vikir} & model & .952 & .333 & -- \\
\midrule
\multicolumn{5}{l}{\emph{Formalize-then-plan (our \detect and $\Phi$)}} \\
Prompt-only & untrained & .086 & .022 & 37.8 / 56.1 \\
Full-translation (NSDL-SFT ckpt) & rules & .513 & .674 & 10.3 / 0.0 \\
Trained-translation & rules & .637 & .406 & 9.4 / 0.0 \\
\midrule
\multicolumn{5}{l}{\emph{NSDL}} \\
NSDL 3B (SFT) & model & .978 & .815 & 0.0 / 0.0 \\
NSDL 7B (SFT) = NSDL-SFT & model & .980 & .970 & 0.0 / 0.0 \\
NSDL 7B (SFT + RL) = NSDL-RL & model & .978 & .970 & 0.0 / 0.0 \\
NSDL 8B, Qwen3-VL (SFT) & model & \textbf{.981} & \textbf{.973} & 0.0 / 0.0 \\
NSDL 7B w/o $\Phi$ (same corpus, SFT) & model & .982 & .871 & 0.0 / 1.9 \\
\bottomrule
\end{tabular}}
\vspace{-0.10cm}
\end{table}

\begin{table}[t]
\caption{LOCUS-Bench accuracy per split, equal-weight mean, and
rejection rate. Judgment and bold as in Table~\ref{tab:viki}; $\Phi$: solver used.}
\label{tab:locus}
\centering
\footnotesize
\setlength{\tabcolsep}{2.5pt}
\resizebox{\columnwidth}{!}{%
\begin{tabular}{llcccccccc}
\toprule
Method & Judgment & $\Phi$ & ID & OOD-S & OOD-M & OOD-X & mean & rej.\% \\
\midrule
\multicolumn{9}{l}{\emph{A: no training on LOCUS}} \\
Qwen2.5-VL-7B zero-shot (our tools) & untrained & \checkmark & .283 & .121 & .135 & .175 & .179 & 22.0 \\
gpt-5.6-terra (our tools) & untrained & \checkmark & .317 & .186 & .120 & .050 & .168 & 42.9 \\
VIKI-R 7B \cite{vikir}, op-level protocol & model & -- & .072 & .014 & .040 & .000 & .032 & 71.3 \\
VIKI-R 7B \cite{vikir}, native protocol & model & -- & .467 & .229 & .255 & .125 & .269 & -- \\
Formalize-then-plan, prompt-only & untrained & \checkmark & .150 & .064 & .100 & .225 & .135 & 31.4 \\
NSDL-SFT trained on VIKI (transfer) & model & \checkmark & .572 & .243 & .350 & .150 & .329 & 2.0 \\
\midrule
\multicolumn{9}{l}{\emph{B: trained on LOCUS, goal verdicts by fixed rules (formalize-then-plan)}} \\
Full-translation (NSDL-SFT ckpt) & rules & \checkmark & .389 & .143 & .060 & .150 & .185 & 6.3 \\
Trained-translation & rules & \checkmark & .689 & .264 & .305 & \textbf{.400} & .415 & 7.0 \\
\midrule
\multicolumn{9}{l}{\emph{C: trained on LOCUS, state judgment in the model}} \\
NSDL 3B (SFT) & model & \checkmark & .722 & .300 & .405 & .350 & .444 & 0.0 \\
NSDL 7B (SFT) = NSDL-SFT & model & \checkmark & .794 & .350 & .420 & .325 & .472 & 0.2 \\
NSDL 7B (SFT + RL) = NSDL-RL & model & \checkmark & \textbf{.839} & \textbf{.429} & \textbf{.460} & .325 & \textbf{.513} & 0.5 \\
NSDL 8B, Qwen3-VL (SFT) & model & \checkmark & .800 & .350 & .435 & .325 & .477 & 0.9 \\
NSDL 7B w/o $\Phi$ (same corpus, SFT) & model & -- & .772 & .257 & .155 & .075 & .315 & 23.2 \\
\midrule
\multicolumn{9}{l}{\emph{D: reference}} \\
GT sketch $\to \Phi$ & ground truth & \checkmark & .983 & 1.000 & .920 & 1.000 & .976 & -- \\
\bottomrule
\end{tabular}}
\vspace{-0.10cm}
\end{table}

On LOCUS-Bench, accuracy follows whether the model was trained to judge
these scenes: a frontier model given the same tools, planners trained
on other data, and our own VIKI-trained checkpoint all stay far below
the LOCUS-trained rows (Table~\ref{tab:locus}). In group A,
\mbox{gpt-5.6-terra} lands on par with the Qwen2.5-VL-7B zero-shot row.
Under its native protocol, VIKI-R 7B is 20.9 points below NSDL-SFT on
the whole set ($p<.0001$), %
and the transfer row, trained on VIKI, also stays far below NSDL-SFT.
Within group C, NSDL-RL has the highest mean: RL adds 4.8 whole-set
points over SFT on LOCUS (46/19, $p=.0011$), %
and Sec.~\ref{sec:exp-judgment} looks at where the gains land. Group D
checks $\Phi$ itself; its imperfect rows are explained in
Appendix~\ref{app:failure}.\looseness=-1

Training the model without the solver brings the structural errors
back and costs accuracy out of distribution on both benchmarks. The
variant without $\Phi$ is trained on the same rows, with the \sched
turns removed and \texttt{[PLAN]} with mechanical ops as the
supervised output. The runtime still compiles the variant's ops into
primitives and a schedule, so the comparison removes the prerequisites
and handoffs the solver adds, and its rejection feedback. On VIKI it matches NSDL-SFT on ID within 0.2 points
but drops 9.9 points on OOD, with the rejection rate rising to
1.9\%; %
on LOCUS the equal-weight mean drops 15.7 points, %
with the rejection rate rising from 0.2\% to 23.2\% and the largest
losses on OOD-M and OOD-X. Of the 63.7 OOD points between NSDL-SFT and
VIKI-R 7B on VIKI-L2, 53.8 appear without the solver, from the corpus
and recipe; the solver adds the remaining 9.9 and removes the
rejections. %
\looseness=-1

When fixed rules or an untrained model make the state judgment,
accuracy falls well below NSDL, and training the model to translate
the scene does not close the gap. Many prompt-only plans are rejected,
and most of the rest are wrong. With fixed-rule verdicts
(full-translation) most rejections disappear, but many plans pass the
validator and are still wrong (VIKI-L2 ID 38.4\%, OOD
32.6\%). %
On VIKI-L2 the trained-translation baseline stays 43.2 points below
NSDL-SFT ($p<.0001$), %
and in 75\% of the OOD rows it loses against full-translation it never
lists a goal (Appendix~\ref{app:failure}): %
the translation task itself still demands judgment, this time in
listing the goals. On LOCUS, retraining recovers 23.2 whole-set points
over full-translation ($p<.0001$), %
but what it repairs is the mismatch of a checkpoint never trained to
translate, and the row stays 9.1 points below NSDL-SFT
($p<.0001$). %
OOD-X is the one split where the trained-translation baseline scores
higher; its gap to NSDL-SFT is three discordant rows ($p=.25$,
Appendix~\ref{app:failure}). %
\looseness=-1

\subsection{The Solver Removes the Structural Errors; the Failures That Remain Are Misjudged States}
\label{sec:exp-failure}

Once the solver writes the structure, the runtime rejects almost
nothing, and the plans that still fail symbolic replay fail because of
what the model asserted about the scene. Runtime rejections fall to
1/560 on LOCUS (NSDL-SFT; 3/560 for NSDL-RL) and 0/3018 on
VIKI. %
Symbolic replay splits the LOCUS NSDL-SFT rows into rejected 1,
simulator-infeasible 109, goal-unmet 143, over-length 18, and correct
289 of 560. %
All 109 infeasible rows come from $\Phi$-accepted schedules and
reflect assertion and enumeration errors of the model in $\sigma$:
takes from closed cabinets (55), hallucinated entities (24),
relay-family assignment errors (29), and one missed activation
intent. %
Crossing the first-\texttt{[GOAL]} verdicts with these classes shows
that 130 of the 143 unmet rows carry a wrong verdict, while 63 of the
109 infeasible rows carry none, placing infeasibility with the asserted
states and assignments rather than with the verdicts. %
Runtime rejection covers only the structural constraints of the action
model of $\Phi$, not whether asserted states, entities, and assignments
match the scene, and the interface of $\Phi$ sets where this boundary
runs: $\sigma$ does not state where an entity is taken from, so a take
from a closed cabinet counts as a judgment error. VIKI shows the same
shape on ID: 30 infeasible, 2 unmet, 4 over-length, and 1764 correct
of 1800, %
the infeasible rows again assertion errors in $\sigma$.\looseness=-1

Paired under the same submission protocol, the variant without $\Phi$
loses on structure while its judgment stays close to NSDL's.
When NSDL-SFT is paired against the variant without $\Phi$ under one
protocol (the transcribed \texttt{[PLAN]} re-expanded, since that
variant has no $\Phi$ output), LOCUS drops 4.6 points (37/11,
$p=.0002$), concentrated in OOD-M, with rejections rising from 102 to
130 rows of 560 (the 102 are transcription slips of NSDL-SFT itself,
rejected only under this protocol, Appendix~\ref{app:training}). %
Under this protocol NSDL-SFT solves 235 of the 560 rows and the variant
209; of the 80-row gap in Table~\ref{tab:locus}, 54 rows come from
submitting the solver's schedule directly and 26 from the structure the
solver writes. %
On VIKI the paired drop is 3.9 points (135/18, $p<.0001$), concentrated
in OOD, with rejections rising from 0 to 23 rows of 3018. %
Per-goal judgment differs only slightly between the two variants:
first-verdict accuracy is .747 against .731 (33/15, $p=.013$), while
candidate-enumeration recall (the share of ground-truth-active goals
the model lists) and verdict accuracy on listed goals are each
indistinguishable ($p\ge.22$), %
so most of the gap between the two variants comes from structure. Structure
should be computed, judgment should be learned, and the deterministic
solver holds the final say over the submitted plan. %
\looseness=-1

\begin{figure}[t]
  \centering
  \includegraphics[width=0.90\columnwidth]{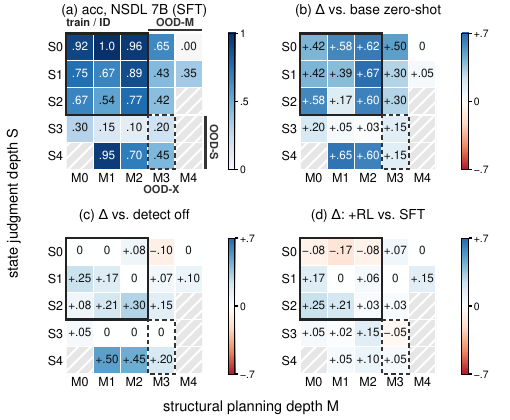}
  \vspace{-0.20cm}
  \caption{\textbf{LOCUS-Bench on the S$\times$M grid (NSDL-SFT).}
  (a) accuracy of the SFT row; (b)--(d) differences within the same
  cell: SFT minus \mbox{Qwen2.5-VL-7B} zero-shot, SFT minus the same
  checkpoint with \detect disabled, and RL minus SFT. Cells pool the
  task families that instantiate them ($n{=}12$--36 ID, 20--60 OOD,
  Appendix~\ref{app:grid}); hatched cells do not exist.}
  \vspace{-0.10cm}
  \label{fig:locus_grid}
\end{figure}

Most goals the model misses are goals it never lists, and their number
does not fall as the evidence improves. NSDL-SFT errs on 230
ground-truth-active goals, of which 91\% are never listed in
\texttt{[GOAL]} at all; %
the never-listed count moves little and non-monotonically across the
four of the evidence rows of Table~\ref{tab:ablation} (detect off
205, NSDL-SFT 209, query scope 243, all entities 227). %
The near-zero S3 (quantifier) row of Fig.~\ref{fig:locus_grid}(c) is
the same mechanism at cell level. A trace audit finds no skip-judged
goal operated on and one of 651 active-judged goals without its op, so
judgment errors reach the submitted plan unrevised. %
\looseness=-1

On the grid, the gains sit in the judgment rows and the remaining
failures in the relay cells and the quantifier row
(Fig.~\ref{fig:locus_grid}). Per-split numbers mix task families and
depths, so the figure places the same samples on the grid and compares
methods within the same cell in panels (b) to (d). %
The zero cell of panel (a) holds 11 of the 29 relay-family assignment
errors plus 9 predicted empty plans. %
In panel (d), the S0 row shows the cost: its ID cells, which pose no
judgment difficulty, move negative under RL. Two boundaries apply:
per-cell accuracy is not monotone along either scale (family
composition differs, and deepening M also raises assignment
difficulty), so LOCUS is not read as a validated two-dimensional
capability scale; %
and the detector is weaker along M (candidate-class macro recall is 11
points below ID), %
so differences along that scale contain a perception share.\looseness=-1

\subsection{Evidence Fixes Only Goals Already in Place}
\label{sec:exp-judgment}

Evidence from \detect is necessary, and ground-truth relations help
only when they are written in the forms the model saw in training
(Table~\ref{tab:ablation}). The table runs on \mbox{NSDL-SFT} and changes
evaluation-time configuration only; the V-ID and V-OOD columns evaluate
the same configuration on VIKI-L2. Block 1 varies the evidence: \detect
off, NSDL-SFT with and without crops, and three rows (GT rel.) that
replace detected relations with ground truth and differ only in scope
and surface form (Appendix~\ref{app:oracle}); the ground-truth rows
carry no crops, so their paired baseline is the crops-disabled row, and,
following \cite{prism}, they serve as controlled probes of the
evidence. The verdict column scores the first \texttt{[GOAL]} block of
each trajectory over 1156 goals (Appendix~\ref{app:oracle}).
Block 2 disables the decoding stop. \detect matters: turning it off
drops 10.4 points ($p<.0001$) with all splits down, %
and the effect carries to VIKI with the same sign and larger size (V
columns, both $p<.0001$). The contribution depends on the in-domain
detector: a zero-shot open-vocabulary replacement (YOLO-World
\cite{yoloworld}, resolution matched) recovers nothing measurable on
either backbone (Appendix~\ref{app:ablations}). %
Yet the first two ground-truth rows are not above NSDL-SFT (query
scope $-3.8$, $p=.011$; %
all entities $-1.4$, not significant, with table mean $+0.8$; against
the crops-disabled row the two differ by $-3.2$ and $-0.9$). %
\looseness=-1

\begin{table}[t]
\caption{Evaluation-time ablations on NSDL-SFT: LOCUS accuracy per
split and mean, verdict accuracy over 1156 goals, VIKI-L2 ID / OOD.}
\label{tab:ablation}
\centering
\scriptsize
\setlength{\tabcolsep}{2.4pt}
\resizebox{\columnwidth}{!}{%
\begin{tabular}{lcccccccc}
\toprule
Configuration & ID & OOD-S & OOD-M & OOD-X & mean & verdict & V-ID & V-OOD \\
\midrule
\multicolumn{9}{l}{\emph{Block 1: evidence}} \\
\detect off & .667 & .207 & .365 & .225 & .366 & .683 & .843 & .327 \\
YOLO, crops (NSDL-SFT) & .794 & .350 & .420 & .325 & .472 & .747 & .980 & .970 \\
YOLO, crops disabled & .794 & .321 & .420 & .350 & .471 & -- & .978 & .970 \\ %
GT rel., query scope & .722 & .307 & .405 & .350 & .446 & .707 & .938 & .758 \\
GT rel., all entities & .728 & .343 & .425 & .425 & .480 & .728 & .937 & .753 \\
GT rel., detector's forms & .806 & .393 & .450 & .325 & .493 & .766 & .985 & 1.000 \\
\midrule
\multicolumn{9}{l}{\emph{Block 2: decoding}} \\
Decoding stop off (LOCUS rej.\ 27.3\%) & .761 & .329 & .155 & .100 & .336 & .621 & .981 & .969 \\
\bottomrule
\end{tabular}}
\vspace{-0.10cm}
\end{table}

The ground-truth rows lose on surface form: facts written in relation
forms never seen in training are misread, and the same facts in the
detector's forms are not. On forms seen in training, the active
targets whose relation is present in the evidence are first-judged
66/66 correct, %
while on unseen forms the model reads located-at-work-area as already
in place. The detector's-forms row scores 6.3 points above the query-scope
row ($p<.0001$) %
and 3.0 points above the crops-disabled row (whole-set .511 to .541,
$p=.002$). %
The V-OOD drop of the first two ground-truth rows is the same
surface-form effect: the two scopes are indistinguishable ($-0.2$
whole-set, $p=.32$), %
which rules out scope, and the ground-truth vocabulary renders most OOD
positions in located-at-area forms that neither training nor the
detector ever produces. The detector's-forms row confirms this on VIKI: it sits above its
crops-disabled baseline by $+0.7$ on ID and $+3.0$ on OOD (both
$p\le.0002$), with every discordant row flipping upward, and its
all-entities variant scores identically on every row. %
On LOCUS its gain comes entirely from goals already in place: first
verdicts on them move from .704 to 1.000, while wrong verdicts on goals
that need action stay at 226 to 228 with and without the relations in
the detector's forms.\looseness=-1 %

Reinforcement learning gains on LOCUS, and its gains are dominated by
goals the model newly lists. RL is reported at the fixed final step
(Appendix~\ref{app:training}), and the gain is 4.8 points on the whole
set (46/19, $p=.0011$; OOD-S $+7.9$, $p=.013$). %
48 of the 65 discordant rows change the \texttt{[GOAL]} verdicts, with
the gains dominated by missed goals newly and correctly listed (26,
against 5 corrected verdicts on listed goals), %
and the rest change the plan under unchanged verdicts. The gain holds
only at the 7B tier, as both 3B RL checkpoints degrade, and a reward
that scores the $\Phi$ output does not help
(Appendix~\ref{app:training}). %
Across the evidence conditions of Table~\ref{tab:ablation}, the
judgment errors persist, and they are reduced mainly by training. %
\looseness=-1

A strong enough open-source teacher matches the default one, and the
decoding stop matters only on the main backbone. With corpus
retraining, swapping the distillation teacher has no significant effect
on LOCUS; on VIKI a Qwen2.5-VL-72B teacher matches the default, and a
\mbox{Qwen3-VL-32B} teacher falls short.
With the decoding stop off, the 7B model answers \texttt{[PLAN]}
without waiting for the tool (rejection rate in the row label of
Table~\ref{tab:ablation}), while Qwen3-VL-8B is unaffected
(Appendices~\ref{app:ablations} and \ref{app:teacher}).\looseness=-1 %

%% file: sections/06_conclusion.tex
\section{CONCLUSION}
\label{sec:conclusion}

For a small VLM that plans for a heterogeneous robot team, the
decisions worth learning are the ones that depend on reading the scene;
the prerequisites, handoffs, and timing that follow can be computed.
Dividing the work this way, NSDL leads the end-to-end and
formalize-then-plan baselines on both benchmarks with models from 3B
to 8B. Computing the structure leaves almost no ordering or scheduling
errors, and most of the goals the model still misses are goals it never
lists. Prior work on formalize-then-plan methods with VLMs places the
bottleneck in reading initial-state relations \cite{vlm-formalizer}. In
our trained setting, ground-truth relations do not reduce the number of
goals the model never lists, so the next step is to teach the model
which goals to list. Future work will also extend NSDL to physical
execution with closed-loop feedback and a larger action
model.\looseness=-1

%% file: sections/07_appendix.tex
\appendices

\section{Evaluation Protocol}
\label{app:protocol}

\textbf{Splits and hygiene.} No evaluation data of either benchmark
flows back into training, rollouts, or checkpoint selection; the LOCUS
ID test split may serve as a training-time monitor, the three OOD
splits are not read until the final report. The reference plans are
at most 15 steps on LOCUS and 11 on VIKI-L2, so the official length
gate $|\bar{\tau}| / |\tau| \ge 0.99$ coincides with $|\tau| \le
|\bar{\tau}|$ on this data.

\textbf{Submission and fallback.} The submitted plan is the last
$\Phi$-accepted schedule. When $\Phi$ never accepts within the tool
budget (27 of 560 LOCUS rows, 1 of 3018 VIKI rows), %
the written \texttt{[PLAN]} is re-expanded: its ops are compiled into
primitives and a schedule without inserting ops, so the plan structure
stays as written. %
The tool budget of 3 executed calls applies to every tool-taking method
except the closed-source baseline, which gets 4, a script setting under
which at most 7 of its rows used the fourth call.

\textbf{Rejection rate.} A row is rejected when any stage of the
runtime raises: the op-level validator (empty plan, missing open before
deposit, cross-robot custody, end-effector capacity exceeded,
capability mismatch, missing or unknown robot, invalid dependency
reference, dependency cycle, and similar codes), the capability check
on the dependency graph, or the step that compiles ops into primitives
and a schedule. Rejected rows score 0 and stay
in the denominator.

\textbf{VIKI-R under two protocols.} The released VIKI-R weights are
run on LOCUS under our op-level protocol (the \texttt{[GOAL]} /
\texttt{[PLAN]} prompt, 524288-pixel cap) and under their native
primitive-schedule protocol (their own system prompt and image
resolution, temperature 0), the latter scored by the official VIKI-L2
scorer with the LOCUS empty-plan extension. Table~\ref{tab:viki} cites the
published VIKI-L2 values; re-running the released VIKI-R weights under
the official protocol reproduces the OOD column within 0.5 points, %
with ID 1.9 (7B) and 4.4 (3B) points lower. %

\section{Failure Decomposition}
\label{app:failure}

Symbolic replay assigns every non-rejected row to exactly one of three
classes in order: simulator-infeasible (a primitive the action checker
refuses, an entity that does not exist, or an unmet temporal
constraint), goal-unmet, and over-length. LOCUS NSDL-SFT: 1
rejected, 109 infeasible, 143 unmet, 18 over-length, 289 correct of
560; %
VIKI NSDL-SFT (ID): 30 infeasible, 2 unmet, 4 over-length, 1764 correct
of 1800. %
The 109 LOCUS infeasible rows break down by the first failing
primitive into takes from a closed cabinet without opening it (55),
entities absent from the scene (24), relay-family assignments that
leave a fixed arm unable to reach or place (29, all at M3--M4), and one
missed activation intent whose plan ends with the temporal chain
unfinished. %
Of the 143 unmet rows, 130 carry a wrong first verdict; of the 109
infeasible rows, 63 carry none. %
Under its native protocol, VIKI-R 7B has 287 simulator-infeasible LOCUS
rows of 560, against 109 for NSDL-SFT. %

The trained-translation baseline loses on VIKI-L2 OOD mostly by never
listing goals, and its lead on LOCUS OOD-X is three rows. On VIKI-L2 it
differs from the full-translation baseline by $-3.4$ points overall
($p=.0004$), split as $+12.4$ on ID and $-26.8$ on OOD (both
$p<.0001$); in 75\% of the OOD rows it loses, it never lists a
goal. %
On LOCUS OOD-X the trained-translation baseline and NSDL-SFT
differ by three discordant rows ($p=.25$, $n=40$), all in the occlusion
family. %
OOD-X excludes the relay cells, where literal translation turns 35 of
the 40 relay rows infeasible. %

The 19 imperfect rows of the reference row of Table~\ref{tab:locus},
which expands ground-truth sketches through $\Phi$, come from how those
sketches were derived and from our re-encoding of the action model.
They are 16 rows outside the training grid whose text-only derivation
assigns the two relay roles in inventory order, landing them reversed,
and 3 ID rows that come from the re-encoding difference: a Move is
demanded before every deposit where the official semantics lets a
fixed arm place in station. %

\section{Ground-Truth Evidence Rows}
\label{app:oracle}

The three ground-truth rows of Table~\ref{tab:ablation} replace the
detector's output with relations computed from the ground-truth state.
Query scope returns relations only for the entities the model asked
about; all entities returns relations for every instruction-relevant
entity, marking invisible ones; the detector's-forms row returns the
same facts written in the relation forms the detector produces in
training (\texttt{on table}, \texttt{in cabinet}, \texttt{on sink},
\texttt{on toaster}) and stays silent on counter, rack, and carrier
positions, which the detector never reports. None of the three returns
crops, so the ground-truth rows are paired with the crops-disabled row
of Table~\ref{tab:ablation}. The verdict column of that table scores
one verdict per goal from the first \texttt{[GOAL]} block of each
trajectory; later blocks, when the model emits any, are not scored, and
the column is not computed for the crops-disabled row.

Given to the trained-translation baseline, ground-truth relations
recover part of its gap to NSDL-SFT: a purpose-trained
relation model (the same backbone fine-tuned to emit placement
relations; two false positives in 560 rows) changes nothing ($-0.2$,
$p=1.0$), while ground-truth relations add $+3.0$ at query scope and
$+5.2$ with all entities (both $p<.0001$), every flipped row flipping
upward; the difference between the two scopes falls in OOD-M, and the
all-entities tier stays below NSDL-SFT on OOD-S ($-7.9$, $p=.035$)
with 28 runtime rejections against 1. %

\section{Training Details}
\label{app:training}

\textbf{RL hyperparameters.} 30 steps; each step samples batches of
192 prompts $\times$ 8 rollouts and keeps 64 prompts whose rollouts
differ in reward (dynamic sampling), so the 30 steps sample about 0.94
epoch of the 6300-prompt LOCUS pool. RL starts from the SFT checkpoint
and is reported at the fixed final step with no evaluation-based
selection;
reward weights $w_T/w_S/w_E = 0.40/0.30/0.15$; protocol penalties of
0.05 per format violation (cap 0.30) and 0.05 per repeated
\sched call (cap 0.15); learning rate $10^{-6}$ without warmup, gradient
clip 1.0, clip ratios 0.2/0.28, KL 0.01, no entropy bonus, token-mean
loss, dynamic sampling, temperature 1.0, response length 3072, prompt
length 4096. %
Both 3B RL checkpoints degrade (LOCUS $-5.4$, $p=.003$; VIKI $-3.8$,
$p<.0001$) and are not listed. %

\textbf{Why the reward scores \texttt{[PLAN]}.} Scoring
\texttt{[PLAN]} scores the expansion when the model copies it
faithfully and penalizes an unfaithful copy, which is itself a model
error. The model writes the expanded plan back as \texttt{[PLAN]}
after $\Phi$ accepts, and under the training supervision
\texttt{[PLAN]} is a transcription of the accepted expansion. Switching the reward to score the $\Phi$
output directly yields no gain on LOCUS ($-2.3$, $p=.09$) and a small
but significant drop on VIKI ($-0.4$, $p=.036$); %
without \texttt{[PLAN]} in the reward the model's copy of the plan
collapses (rescored on the transcribed \texttt{[PLAN]}, VIKI falls to
.792/.597). %

\textbf{Same-protocol pairing.} The paired comparison of
\mbox{Sec.~\ref{sec:exp-failure}} re-expands the transcribed \texttt{[PLAN]}
of both variants, since the variant without $\Phi$ has no $\Phi$
output. Under
this protocol 102 LOCUS rows of NSDL-SFT are rejected; they are
transcription slips of NSDL-SFT itself (the submitted
$\Phi$-accepted schedules of the same rows pass) and are rejected only
under this protocol. Under this protocol NSDL-SFT solves 235 of the 560
LOCUS rows and the variant without $\Phi$ solves 209
(Sec.~\ref{sec:exp-failure}). %

\section{Ablation Numbers Cited in the Text}
\label{app:ablations}

Table~\ref{tab:cited} lists the ablation numbers that
Sec.~\ref{sec:exp} and Appendices~\ref{app:failure} to
\ref{app:teacher} cite without a row in Tables~\ref{tab:viki} to
\ref{tab:ablation}, plus whole-set values of four Table~\ref{tab:ablation}
rows. Each comes with the paired baseline it is compared against, the
difference, and the $p$ value of the paired test where one was run.

\begin{table*}[t]
\caption{Ablation numbers cited in Sec.~\ref{sec:exp} and Appendices~\ref{app:failure} to \ref{app:teacher}, with whole-set values
of Table~\ref{tab:ablation} rows. LOCUS: whole set; VIKI: ID / OOD.}
\label{tab:cited}
\centering
\scriptsize
\setlength{\tabcolsep}{6pt}
{%
\begin{tabular}{lll}
\toprule
Configuration & Value & Paired baseline, $\Delta$, $p$ \\
\midrule
NSDL-SFT, crops disabled (LOCUS) & .511 & NSDL-SFT with crops, $-0.5$, $p=.61$ \\
GT rel., detector's forms (LOCUS) & .541 & crops disabled, $+3.0$, $p=.002$ \\
GT rel., detector's forms, all entities (VIKI) & same as query scope & on every row \\
GT rel., detector's forms (VIKI) & .985 / 1.000 & crops disabled, $+0.7$ / $+3.0$, $p\le.0002$ \\
YOLO-World replacing the detector, 7B & no detector $+0.0$ & $p=1.0$ \\
YOLO-World replacing the detector, 8B & no detector $+0.9$ & $p=.44$ \\
Trained-translation $+$ relation model (LOCUS) & $-0.2$ vs.\ detector & $p=1.0$ \\
Trained-translation $+$ GT rel., query scope (LOCUS) & $+3.0$ vs.\ detector & $p<.0001$, all flips upward \\
Trained-translation $+$ GT rel., all entities (LOCUS) & $+5.2$ vs.\ detector & $p<.0001$; OOD-S $-7.9$ vs.\ NSDL-SFT \\
Reward scoring the $\Phi$ output (LOCUS / VIKI) & $-2.3$ / $-0.4$ & NSDL-RL, $p=.09$ / $p=.036$ \\
Reward without \texttt{[PLAN]} (VIKI, rescored) & .792 / .597 & \\
3B SFT$+$RL (LOCUS / VIKI) & $-5.4$ / $-3.8$ & 3B SFT, $p=.003$ / $p<.0001$ \\
Teacher Qwen2.5-VL-72B (VIKI OOD) & .958 & default-teacher control .967 \\
Teacher Qwen3-VL-32B (VIKI OOD) & .808 & \\
Decoding stop off, Qwen3-VL-8B & byte-identical outputs & \\
Decoding stop off, 7B (VIKI) & within 0.1 & NSDL-SFT \\
VIKI-R 7B native, infeasible rows (LOCUS) & 287 / 560 & NSDL-SFT 109 / 560 \\
Detector macro recall along M & 11 points below ID & \\
\bottomrule
\end{tabular}}
\end{table*}

\section{Teacher and Decoding Stop}
\label{app:teacher}

The choice of teacher matters only for the Qwen3-VL-32B teacher, and
the decoding stop matters only for the 7B model. Think segments of the
SFT corpus are synthesized by \mbox{gpt-5.6-terra}. Swapping the teacher and retraining the corpus gives, on VIKI-L2 OOD,
.958 under a Qwen2.5-VL-72B teacher against .967 for the paired
default-teacher control, and .808 under a Qwen3-VL-32B
teacher; %
on LOCUS the change is not significant. The decoding stop truncates
generation after an issued tool call; with it off, the main backbone
answers \texttt{[PLAN]} before the tool returns on 27.3\% of LOCUS
rows (rejected), while Qwen3-VL-8B produces byte-identical outputs and
VIKI changes within 0.1 points. %

\section{Grid Statistics}
\label{app:grid}

Table~\ref{tab:locus_stats} gives the size and the reference-plan
lengths of each split. Per-cell sample counts in Fig.~\ref{fig:locus_grid} are 6 per
family-cell on the ID split and 20 per family-cell on the OOD splits;
cells pool the family-cells that instantiate them, giving 12--36
samples per ID cell and 20--60 per OOD cell.

\begin{table}[!htbp]
\caption{LOCUS-Bench splits. Reference-plan length: mean over all
plans / mean over non-empty plans / maximum.}
\label{tab:locus_stats}
\centering
\footnotesize
\setlength{\tabcolsep}{3pt}
\begin{tabular}{lrrccrccc}
\toprule
Split & $n$ & family-cells & S & M & empty & mean & non-empty & max \\
\midrule
train & 6300 & 30 & 0--2 & 0--2 & 972 & 5.1 & 6.0 & 11 \\
ID test & 180 & 30 & 0--2 & 0--2 & 31 & 5.0 & 6.1 & 11 \\
OOD-S & 140 & 7 & 3--4 & 0--2 & 12 & 6.5 & 7.1 & 15 \\
OOD-M & 200 & 10 & 0--2 & 3--4 & 27 & 6.3 & 7.3 & 10 \\
OOD-X & 40 & 2 & 3--4 & 3 & 3 & 7.5 & 8.1 & 10 \\
\bottomrule
\end{tabular}
\end{table}